\documentclass{article}

\pdfoutput=1

\usepackage[utf8]{inputenc}
\usepackage{graphicx}
\usepackage{subcaption}
\usepackage{textcomp,gensymb}

\usepackage[final, nonatbib]{neurips_2020}
\usepackage[square,sort,comma,numbers]{natbib}

\title{Modeling Social Interaction for Baby in Simulated Environment for Developmental Robotics}

\author{
  Md Ashaduzzaman Rubel Mondol \And Aishwarya Pothula \And Deokgun Park \And \\
  Computer Science and Engineering\\
  University of Texas at Arlington\\
  Arlington, Texas USA \\
  \texttt{ \{mdashaduzzaman.mondol, aishwarya.pothula\}@mavs.uta.edu, deokgun.park@uta.edu} \\
}

\date{October 2020}

\begin{document}

\maketitle

\begin{abstract}
    Task-specific AI agents are showing remarkable performance across different domains. But modeling generalized AI agents like human intelligence will require more than current datasets or only reward-based environments that don't include experiences that an infant gathers throughout its initial stages. In this paper, we present Simulated Environment for Developmental Robotics (SEDRo). It simulates the environments for a baby agent that a human baby experiences throughout the pre-born fetus stage to post-birth 12 months. SEDRo also includes a mother character to provide social interaction with the agent. To evaluate different developmental milestones of the agent, SEDRo incorporates some experiments from developmental psychology.
\end{abstract}

\pdfoutput=1
\section{Introduction}
To develop an Artificial Intelligence agent that can perform diverse tasks is still far from reality. Although task-specific AI agents are outperforming humans in many fields, there is none to perform all the tasks a single human can do. To build such an agent, combining different current task-specific models poses a great challenge as training time and dataset requirements grow exponentially as the number of tasks grows.

A human child is born with no experience of the world but over time learns to do many tasks that require complex sequential steps. The brain follows a universal algorithm to perform all the diverse tasks. To build a single agent to perform diversified tasks, we have to find out such kind of universal learning mechanism. Many researchers use a physical robot to study and test such a mechanism~\cite{johansson2020epi, metta2008icub, gouaillier2009mechatronic}. Building such an agent would require an environment that facilitates the necessary elements for longitudinal learning. Training that kind of agent in the real world is costly in terms of both time and money. Moreover, it may not be possible to reproduce all the scenarios. Computer simulated environments that can provide realistic experiences has become a common approach to diminish this problem.

The surrounding environment plays an important role in an infant's learning. Studies suggest that social interaction influences cognitive development. Right from birth, the infant's social interaction begins with its family members. The motherese or Infant Directed Speech (IDS) has a significant impact on the infants' cognitive development and language acquisition\citep{catherine2013MothereseInteraction}. And mostly the mother plays a vital role.

Many researchers have been using computer-simulated environments to decode the different abilities of human baby like vision\citep{fuke2007facePerceptionSimulation}, motor skills\citep{savastano2012reachingSimulation} or in modeling Curiosity, Intrinsic Motivation \citep{barto2004IMSimulation, Fiore2008ratSimulationIM}. There have been some simulated robot platforms also like iCub \citep{Tikhanoff2012iCub}, Webot \citep{Michel2004Webots} including some works of Fetus environment\citep{kuniyoshi2007fetusSimulator, kuniyoshi2010fetusSimulator}. Simulated computer games like VizDoom\citep{kempka2016VizDoom}, Obstacle Tower Challenge\citep{juliani2019obstacleTower}. But these platforms don't provide the environments a newborn baby experiences throughout its first year including the social interaction with a mother or other family members.

\begin{figure}
    \centering
    \begin{subfigure}[b]{.24\textwidth}
        \centering
        \includegraphics[width = 1\textwidth]{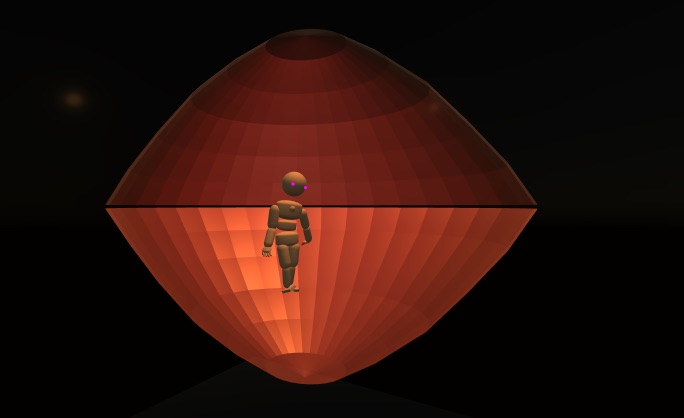}
        \caption{}
        \label{fig:fetus_env}
    \end{subfigure}
    \hskip1mm
    \begin{subfigure}[b]{.24\textwidth}
        \centering
        \includegraphics[width = 1\textwidth]{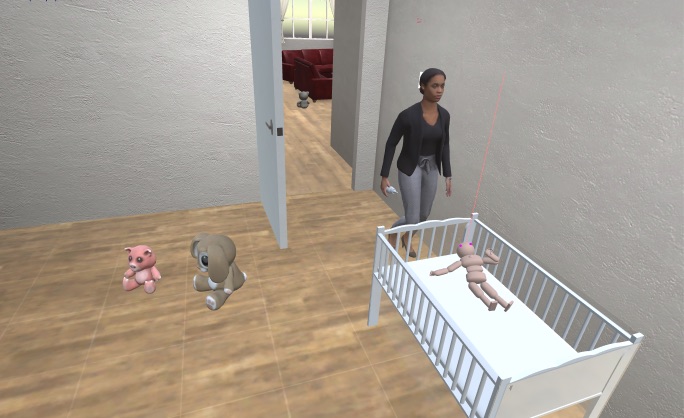}
        \caption{}
        \label{fig:after_birth_env}
    \end{subfigure}
    \begin{subfigure}[b]{.24\textwidth}
        \centering
        \includegraphics[scale=0.09]{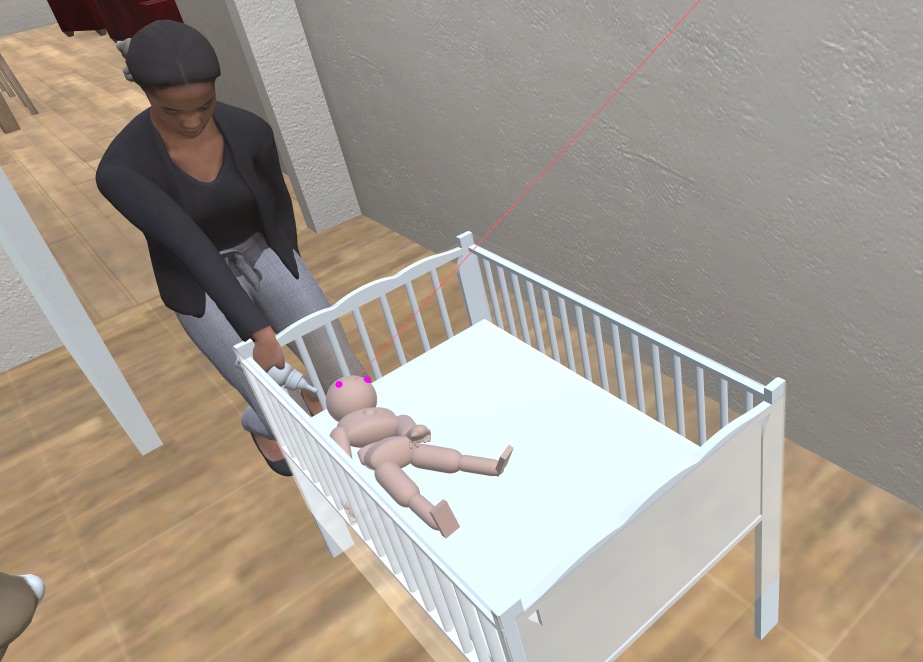}
        \caption{}
        \label{fig:mother_feeding}
    \end{subfigure}
    \begin{subfigure}[b]{.24\textwidth}
        \centering
        \includegraphics[scale=0.115]{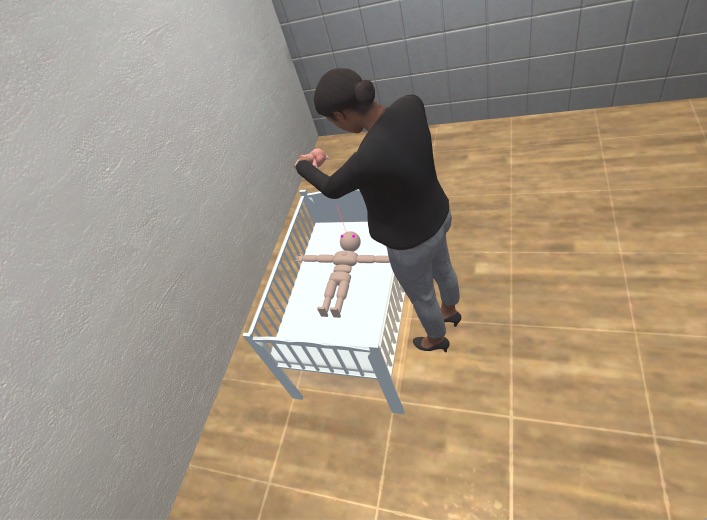}
        \caption{}
        \label{fig:mother_toy}
    \end{subfigure}
    \caption{\small Screenshots from SEDRo environments. (a) Fetus environment. Almost dark space with no visual capability of the baby. (b) After birth house environment with a Mother character and some other toys. (c) Mother is feeding the baby. (d) Mother is showing a toy to the baby}
    \label{fig:sedro}
\end{figure}

In this paper, we present our ongoing work on building a Simulated Environment for Developmental Robotics (SEDRo) to facilitate the development of Generalized Intelligence of baby agent~\cite{pothula2020sedro}. A mother character is added to interact with the baby agent. There will be different stages based on different ages of an infant after the birth as well as one stage that simulates the womb. Each stage will require the learning of previous stages to support incremental developments. Figure~\ref{fig:sedro} shows the screenshots of the SEDRo.
\section{Proposed Environment}
SEDRo will simulate the minimal experience of a baby starting from the fetus stage to 12 months after birth. The key part of SEDRo will be a body of the baby agent and a surrounding environment for the agent to interact. Another important part will be a care-giving character, which will interact with the agent as part of social interaction. Also, there will be other interactive objects like furniture, toys, etc. The agent may interact with the surrounding objects in the room. A model of the agent can interact with the environment with the interface which is an extension of the OpenAI Gym API~\cite{brockman2016openai}. There will be four developmental stages with two different environments (Fetus and After-birth) to mimic different stages of the baby- 1) Fetus, 2) Immobile, 3) Crawling, and 4) Walking stage. Each one will provide a different experience for the agent and also unfold the new capabilities of the agent.

\subsection{The Agent}
The agent body has been developed with capabilities to crawl, walk, grasp objects, and follow the mother's attention. But these capabilities will unfold gradually in each subsequent stage. Initially, the maximum torque value for the joint motion is too small to support walking, but they will increase over time to enable it.   The agent has 64 degree of freedom for the body part movements.

\begin{figure}[ht]
    \centering
    \begin{subfigure}[b]{.23\textwidth}
        \centering
        \includegraphics[width = 1\textwidth]{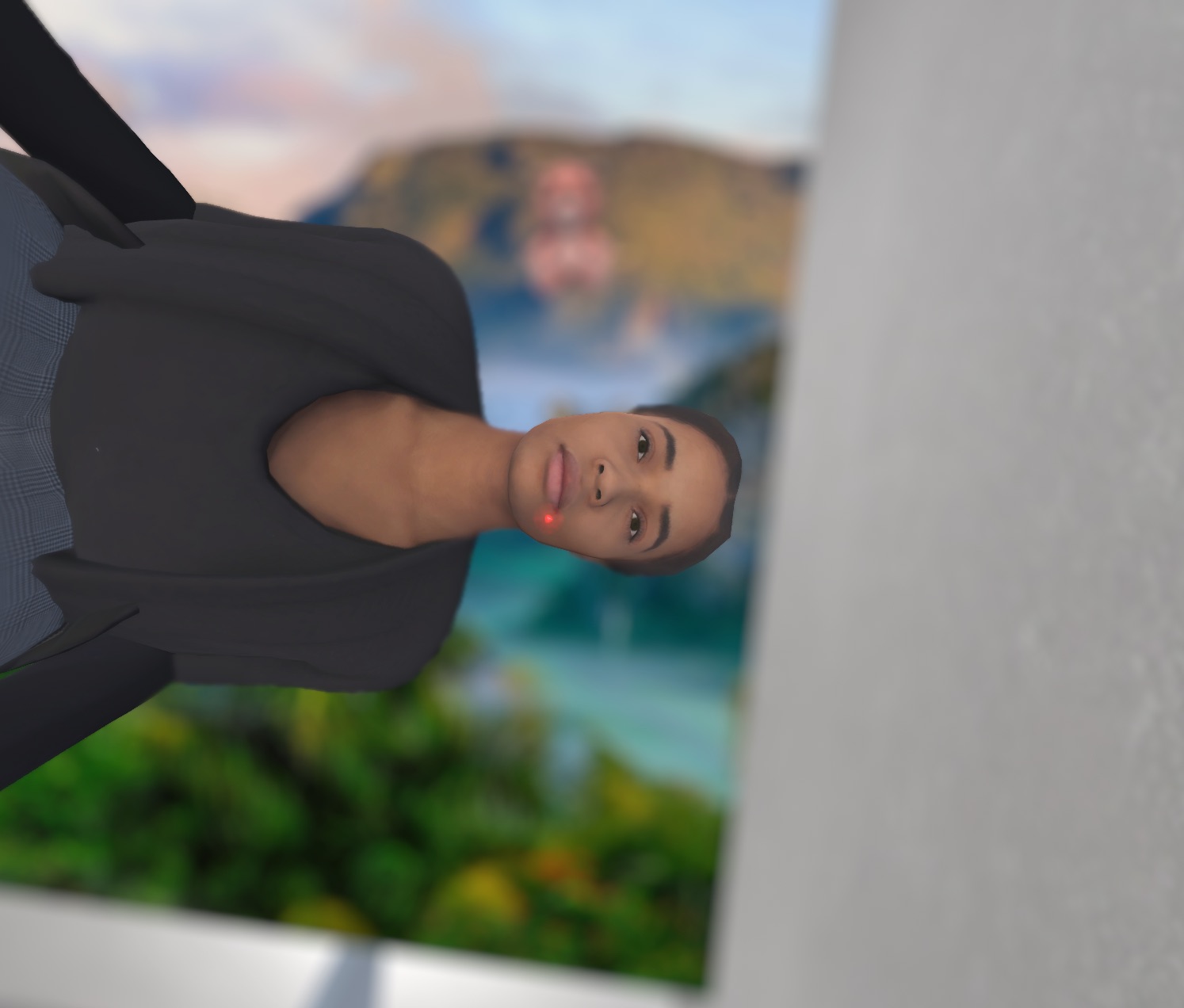}
        \caption{}
        \label{fig:agent_vision_near_peri}
    \end{subfigure}
    \hskip1mm
    \begin{subfigure}[b]{.23\textwidth}
        \centering
        \includegraphics[width = 1\textwidth]{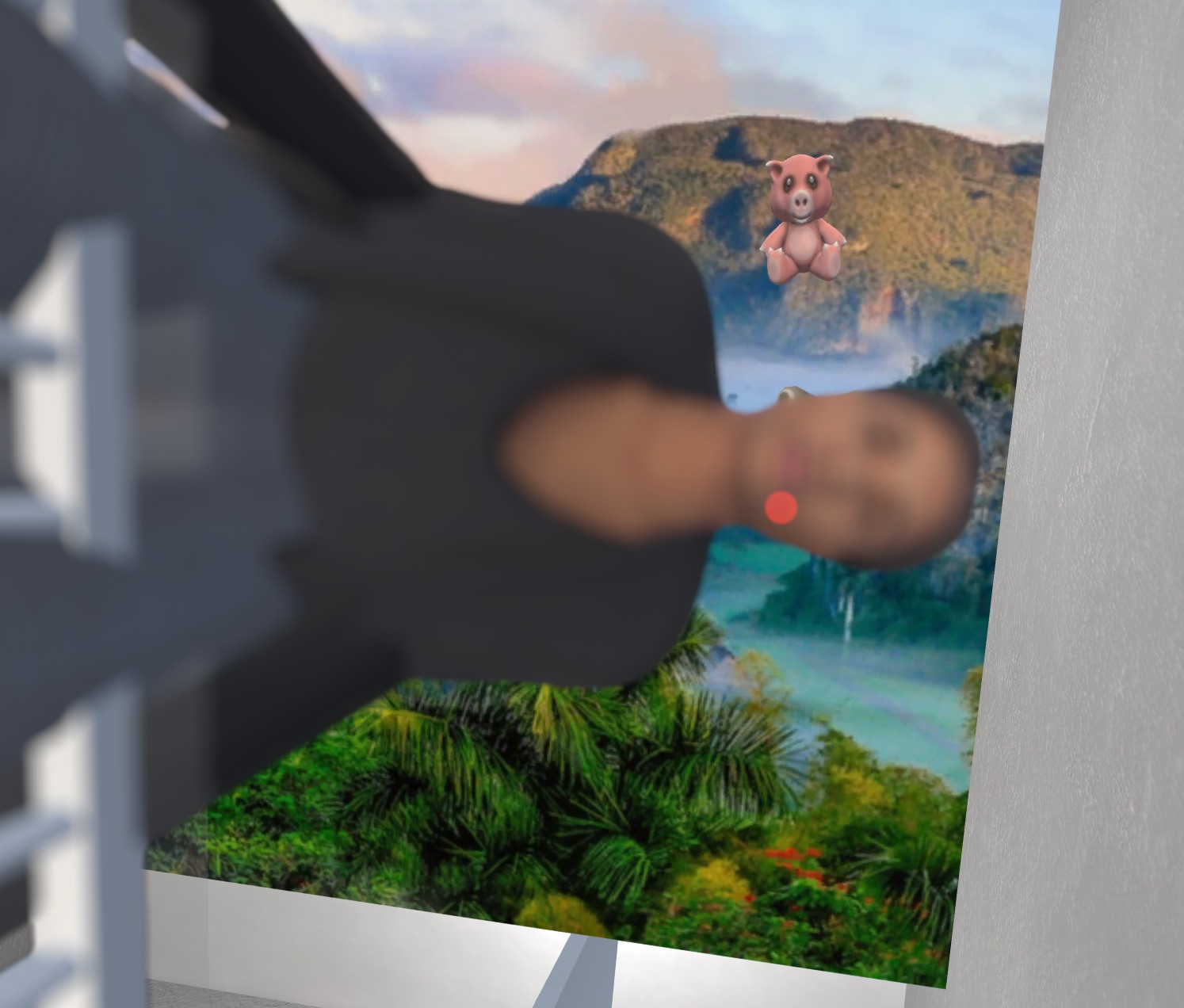}
        \caption{}
        \label{fig:agent_vision_far_peri}
    \end{subfigure}
    \hskip1mm
    \begin{subfigure}[b]{.23\textwidth}
        \centering
        \includegraphics[width = 1\textwidth]{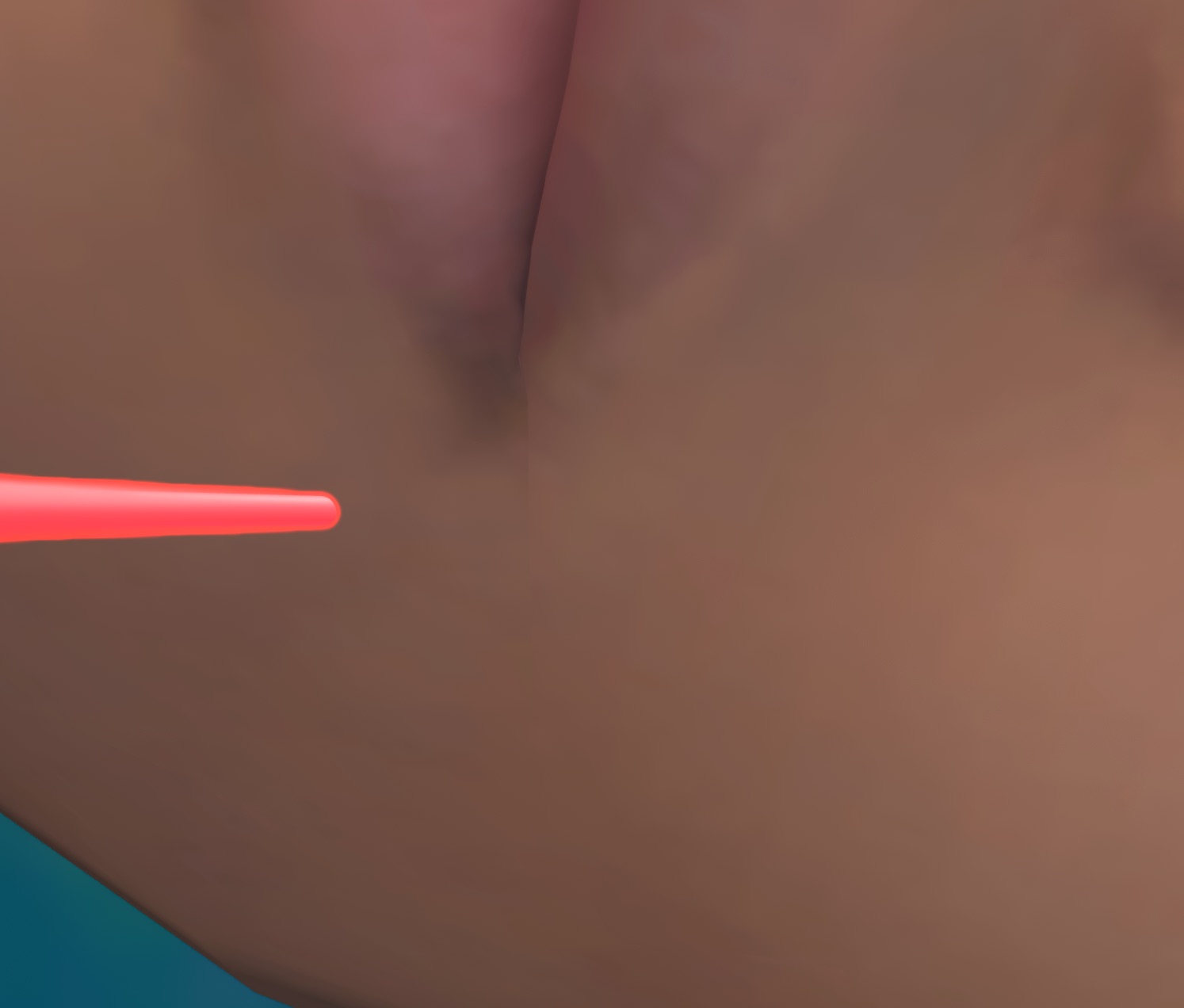}
        \caption{}
        \label{fig:agent_vision_in_focus}
    \end{subfigure}
    \hskip1mm
    \begin{subfigure}[b]{.23\textwidth}
        \centering
        \includegraphics[width = 1\textwidth]{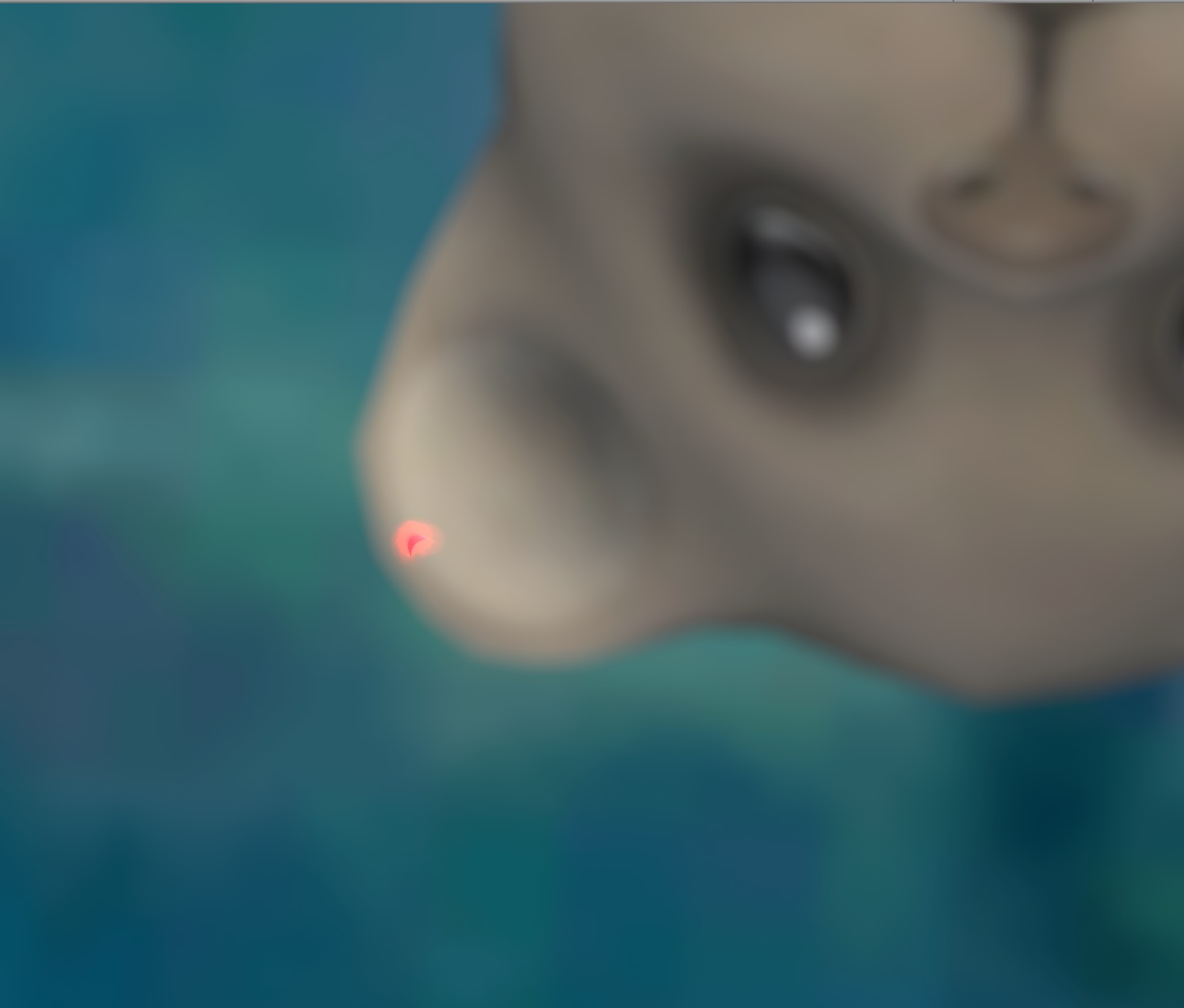}
        \caption{}
        \label{fig:agent_vision_out_of_focus}
    \end{subfigure}
    
    \begin{subfigure}[b]{.23\textwidth}
        \centering
        \includegraphics[width = 1\textwidth]{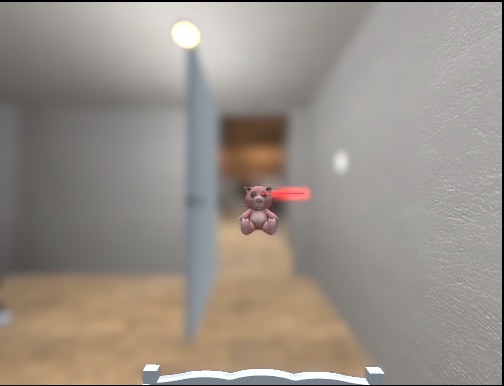}
        \caption{}
        \label{fig:agent_vision_left_peri}
    \end{subfigure}
    \hskip1mm
    \begin{subfigure}[b]{.23\textwidth}
        \centering
        \includegraphics[width = 1\textwidth]{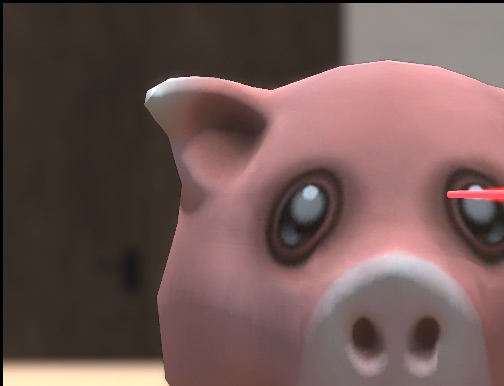}
        \caption{}
        \label{fig:agent_vision_left_central}
    \end{subfigure}
    \hskip1mm
    \begin{subfigure}[b]{.23\textwidth}
        \centering
        \includegraphics[width = 1\textwidth]{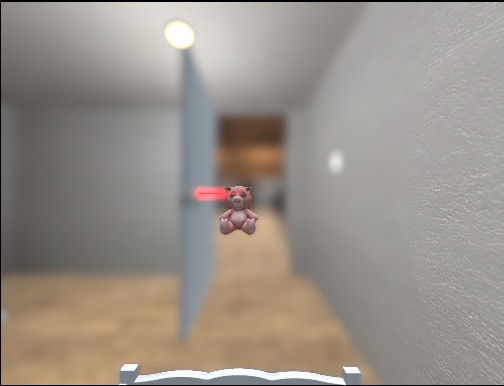}
        \caption{}
        \label{fig:agent_vision_right_peripheral}
    \end{subfigure}
    \hskip1mm
    \begin{subfigure}[b]{.23\textwidth}
        \centering
        \includegraphics[width = 1\textwidth]{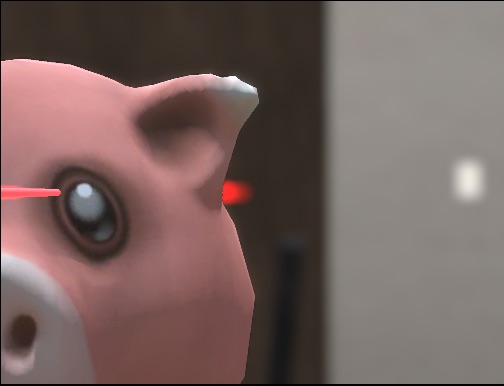}
        \caption{}
        \label{fig:agent_vision_right_central}
    \end{subfigure}
    \caption{\small Baby Agent's vision. (a) Nearsighted peripheral vision. (b) Farsighted peripheral vision. (c) Central vision in-focus. (d) central vision out-of-focus. (e) Left peripheral vision. (f) Left central vision. (g) Right peripheral vision. (h) Right central vision}
    \label{fig:agent_vision}
\end{figure}

\paragraph{Vision}
The agent has a binocular vision system with two eyes. Both eyes have a combined 3 degree of freedom; 1 for horizontal, 1 for vertical rotation of the eyeballs, and 1 to adjust focus. Each eye contains two cameras to simulate the central(8\degree) and peripheral(100\degree) vision of the human eye. There is one more camera on the head placed between the eyes of the agent to generate the combined visual image of both eyes. This is an optional camera provided for debugging purposes. Also, the depth of field effect has been applied to all the cameras for nearsighted focusing effects, since early infants cannot focus beyond arm's length. Figure \ref{fig:agent_vision} shows different visual inputs for the agent.

\paragraph{Tactile Sensitivity}
As part of tactile sensitivity, the agent is equipped with touch sensors of varying density, since the perception of touch differs across the body ~\citep{rochelle2014TouchDensity}. There are a total of 2110 tactile sensors placed across the body.  About half of the sensors are placed on the head. The sensors detect the touch, based on the collision detection mechanism. Each sensor generates a value of 1 when a 'touch' occurs or 0 otherwise. A sparse status vector is generated consisting of all sensor status and sent as part of observations.

\paragraph{Proprioception}
To learn the association of spatial locations and body parts movements, the baby will require its current joint positions along with visual information. In SEDRo, the current positions and rotations of all the body joints are given to the agent's observations. In total, 469 observations will be provided with continuous values ranging from -1 to 1. This vector will also include each joint's velocity and angular velocity to help understand the current body movements.

\paragraph{Interoception}
The baby's stomach food level is also given with the observation vector as a body's internal sensitivity. It represents the current percentage of food available in the stomach. With time, the food level reduces. When the food level falls below a certain threshold representing the hunger, the baby needs to cry to get food. Whenever the baby cries, the mother will feed it and the stomach food level will increase.
\pdfoutput=1
\subsection{Modeling the Motherese}

We implemented a mother character for interacting with the baby. Both the baby and mother are inside a small house. The baby is placed in the crib or on the floor based on different situations and the baby's age.

\paragraph{Building the Mother}
While facilitating the baby agent's intelligence development, one challenging part is building a mother that can interact with the baby.  We tackle this challenge by limiting the experience up to the first year after birth because most interaction in this period does not require open-ended back-and-forth interaction.   
The mother has been programmed with some pre-programmed action capabilities to interact with the baby. 
We are building a library of mother's actions based on real-life mother-child interaction. 
Currently, we are manually building scenarios and we plan to analyze different video recordings from the houses with newborn babies. 
To make realistic behaviors, we will create the movements of the mother with pre-recorded motion captured(Mocap) animations based on those real footages. So, we will get a library with various responses of the mother for the same type of baby's actions.

\paragraph{Interaction with baby}
As a concrete example of the social scenario, feeding the baby will be the mother's regular interaction. The mother will feed the baby at pre-scheduled times of the day. Also, the baby will cry when the stomach food level is below the threshold and the mother will begin feeding scenario.  When feeding, it will move towards the baby and start the feeding. While walking and feeding, the mother can avoid different obstacles and also adjust body positions during the feeding animation based on the baby's current location.

Providing Infant Directed Speech (IDS) is another significant role played by mothers that helps with a child's development. For IDS in SEDRo, the mother character will be talking to the baby with small words and physical expressions like nodding head while looking at the baby, moving arms. One limitation for implementing the mother's vocal is that sound can't be added directly as part of observation for the agent model yet. To overcome this limitation, in our initial version of SEDRo, we will use a one-hot encoded vector of length 26 to represent one English character at every time frame.

For joint attention, the mother will be holding different objects e.g. toys in front of the baby, looking at it, and describe its identity in small words. Also at a more developed stage, if the baby grabs or touches any object, the mother will describe it. Figure. \ref{fig:mother_feeding}, \ref{fig:mother_toy} shows mother's interaction with baby.

\section{Evaluation of Development}
\begin{figure}[t]
    \centering
    \begin{subfigure}[b]{.49\textwidth}
        \centering
        \includegraphics[width = 1\textwidth]{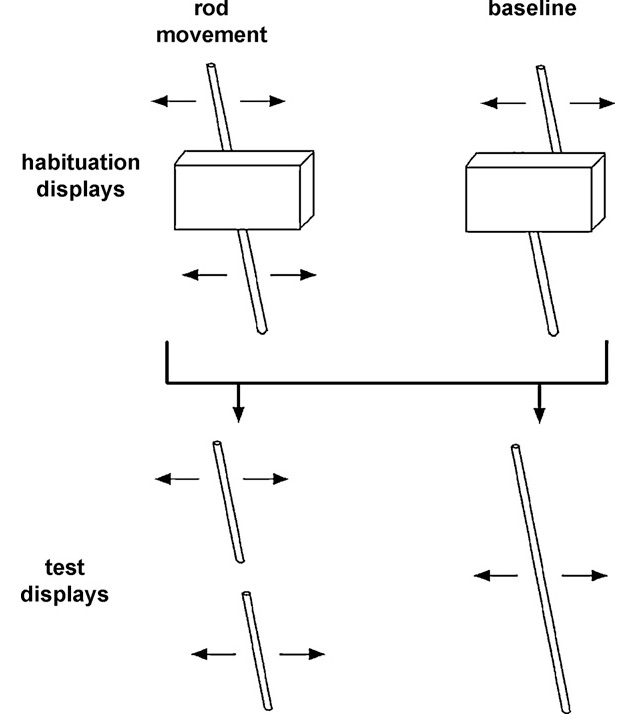}
        \caption{}
        \label{fig:paper_rod_exp}
    \end{subfigure}
    \hskip1mm
    \begin{subfigure}[b]{.49\textwidth}
        \centering
        \includegraphics[width = 1\textwidth]{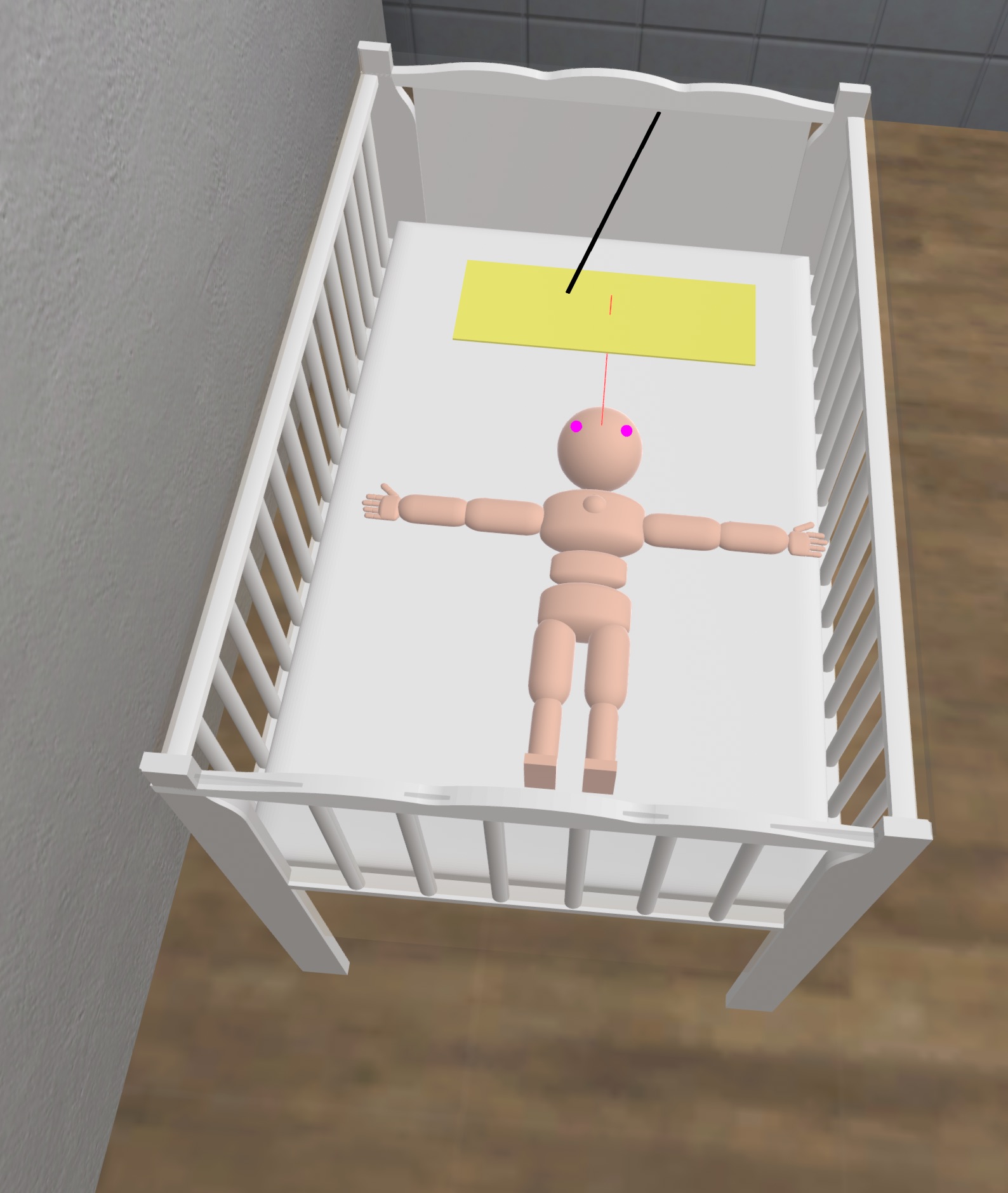}
        \caption{}
        \label{fig:paper_rod_sedro}
    \end{subfigure}
    \caption{\small Evaluation experiments. (a) Paper rod experiment to evaluate unity perception\citep{slater1990evaluation}. (b) Paper rod experiment simulation in SEDRo}
    \label{fig:evaluation}
\end{figure}

There have been many experiments to evaluate and track the progress of cognitive, motor skill, and visual developments of human babies in Developmental Psychologies \cite{cangelosi2015developmental}. Similarly, SEDRo will provide different experiments to evaluate the developmental progress of the agent. Figure \ref{fig:paper_rod_exp} shows one such experiment where a moving rod occluded by a box is shown to the baby. Newborn babies under three months age think of it as two different rods but, older babies see it as a single rod\citep{slater1990evaluation}. This test is to evaluate the unity perception of babies. Similarly, more experiments will be provided in SEDRo from developmental psychologies.

\section{Discussion}
So far, we have presented our in-progress works. SEDRo is currently being implemented using the Unity 3D game engine. It will be improved further over time as we add more social interaction scenarios between the mother character and the agent. 

In this version, we have simulated the motherese voice with character sequence observations. This can be further improved by adding the audio data directly to the observations. Representing voice as a text has a limitation. It's not possible to add variations in a speech this way, but generating prosodic voice requires variations in tone and length of the sounds. And a prosodic voice from motherese plays an important role in language development for the babies\citep{catherine2013MothereseInteraction}. In the future version, we plan to include vocal audio data as observation for the agent model.

\newpage
\bibliographystyle{abbrv}
\bibliography{references.bib}

\end{document}